\documentclass[conference]{IEEEtran}
\IEEEoverridecommandlockouts
\usepackage{amsmath,amssymb,amsfonts}
\usepackage{graphicx}
\usepackage{textcomp}
\usepackage[dvipsnames]{xcolor}
\usepackage{subfiles} 
\def\BibTeX{{\rm B\kern-.05em{\sc i\kern-.025em b}\kern-.08em
    T\kern-.1667em\lower.7ex\hbox{E}\kern-.125emX}}

\usepackage{biblatex} 
\usepackage{mathtools, amssymb, amsthm}
\usepackage{pgfplots}
\usepackage{multirow}
\usepackage{booktabs}
\usepackage{graphicx}
\usepackage{pgfplotstable}
\usepackage{csvsimple}
\usepackage{hhline}
\usepackage[printonlyused,withpage]{acronym}
\usepackage{tabularx}
\usepackage[figuresright]{rotating}
\usepackage{makecell}
\usepackage{siunitx,etoolbox}
\usepackage[nomessages]{fp}
\pgfplotsset{compat=1.18}
\usepgfplotslibrary{fillbetween}
\usepackage{caption}
\usepackage{subcaption}
\usepgfplotslibrary{groupplots}
\usepackage{amsmath}
\usepackage{rotating}
\usepackage{placeins}
\usepackage[export]{adjustbox}
\usepackage{algpseudocode}
\usepackage[linesnumbered]{algorithm2e}
\usepackage{placeins}
\usepackage{afterpage}
\usepackage{amsmath}
\usepackage{hyperref}

\definecolor{random-search}{named}{gray}
\definecolor{unaware}{named}{BrickRed}
\definecolor{wordtype-aware}{named}{OrangeRed}
\definecolor{word-aware}{named}{YellowGreen}
\definecolor{llama}{named}{blue}
\definecolor{falcon}{named}{YellowOrange}
\definecolor{qwen}{named}{Purple}

\colorlet{replaced}{red!100}
\colorlet{inserted}{Green!100}

\let\oldtwocolumn\twocolumn
\renewcommand\twocolumn[1][]{%
    \oldtwocolumn[{#1}{
        \begin{center}
            \includegraphics[width=\textwidth]{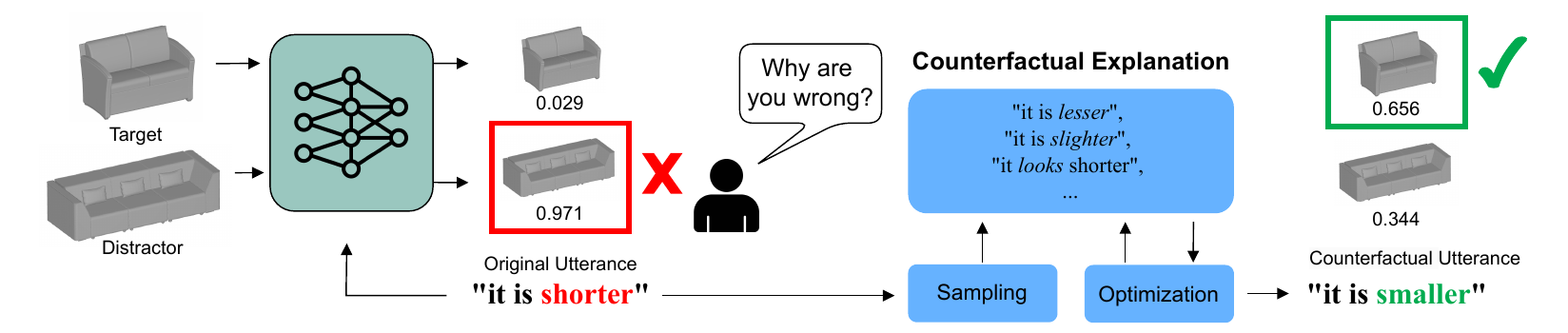}
            \captionof{figure}{Visual illustration of our method for explaining misclassified samples in object referent identification---the task of selecting a specific target object given a natural language description \cite{mitra2024whichone}. When a model incorrectly identifies an object despite being provided with an accurate description, the question arises ``Why is the model wrong?". Our method answers this question by generating counterfactual utterances that show minimal modifications required for a correct prediction.}
            \label{fig:overview}
        \end{center}
    }]
}
    
\begin{document}

\title{Why Are You Wrong? Counterfactual Explanations for Language Grounding with 3D Objects 
\\



}

\DeclareRobustCommand*{\IEEEauthorrefmark}[1]{%
  \raisebox{0pt}[0pt][0pt]{\textsuperscript{\scriptsize #1}}%
}

\author{\IEEEauthorblockA{Tobias Preintner\IEEEauthorrefmark{1,2},
Weixuan Yuan\IEEEauthorrefmark{2},
Qi Huang\IEEEauthorrefmark{1}, 
Adrian König\IEEEauthorrefmark{2}, 
Thomas Bäck\IEEEauthorrefmark{1},
Elena Raponi\IEEEauthorrefmark{1} and
Niki van Stein\IEEEauthorrefmark{1}}
\IEEEauthorblockA{\IEEEauthorrefmark{1}Institute of Advanced Computer Science, Leiden University, Leiden, The Netherlands}
\IEEEauthorblockA{\IEEEauthorrefmark{2}BMW Group, Munich, Germany}}%

\maketitle


\begin{abstract}
Combining natural language and geometric shapes is an emerging research area with multiple applications in robotics and language-assisted design. A crucial task in this domain is object referent identification, which involves selecting a 3D object given a textual description of the target. Variability in language descriptions and spatial relationships of 3D objects makes this a complex task, increasing the need to better understand the behavior of neural network models in this domain. However, limited research has been conducted in this area. Specifically, when a model makes an incorrect prediction despite being provided with a seemingly correct object description, practitioners are left wondering: ``Why is the model wrong?". In this work, we present a method answering this question by generating counterfactual examples. 
Our method takes a misclassified sample, which includes two objects and a text description, and generates an alternative yet similar formulation that would have resulted in a correct prediction by the model. 
We have evaluated our approach with data from the ShapeTalk dataset along with three distinct models. Our counterfactual examples maintain the structure of the original description, are semantically similar and meaningful. They reveal weaknesses in the description, model bias and enhance the understanding of the models behavior. Theses insights help practitioners to better interact with systems as well as engineers to improve models.

\end{abstract}


\section{Introduction}
Jointly combining natural language and geometric shapes is an emerging area of research \cite{sun2024da4lg}. Grounding language with geometric shapes is of great interest in the fields of robotics \cite{thomason2022snare} and language-assisted design \cite{achlioptas2023shapetalk, slim2024shapewalk}. For example, a human can interact with a robot via language, describing objects, or a designer can give text prompts on how to modify a geometry. 
 People use geometric and physical properties of objects when describing them \cite{thomason2022snare}. 
For example, asking a robot to bring the cup with the wide handle requires the robot not only to identify cups, but also to differentiate between multiple cup options based solely on geometric properties, such as the type of handle.
This problem is also known as \textit{object referent identification}, which is the task of selecting a specific object given a natural language description \cite{mitra2024whichone}. 

Despite the fundamental importance of this task for language grounding with 3D objects, little work has been done in adding understanding and explainability. In particular, language interactions via short, context-dependent prompts, linguistically known as \textit{utterances}, is highly relevant, as it represents the crucial intersection between the human and the neural network.
Thus, if a robot selects the incorrect cup in the previous example, despite the human having correctly described the object, two questions arise, ``Why are you wrong?'' and ``How could I have better described the object". An intuitive method to answer this question is to provide an alternative, similar utterance that points out what exactly must have changed in the formulation, such that the robot would have picked the right cup. This is also known as a \textit{counterfactual explanation}, which is an example-based explainable artificial intelligence (XAI) technique. Counterfactual explanations make prediction models more explainable by providing hypothetical examples that lead to the desired output \cite{verma2020counterfactual}.
In this paper, we present a method that generates valid and meaningful counterfactual utterances for object referent identification with 3D objects.

Our method first formulates alternative utterances using a sampler, followed by an optimization process towards the correct prediction, while keeping the semantic meaning close to the original utterance. The generated counterfactual utterances reveal weaknesses in the object description, model bias towards specific words, and enhance the understanding of the models behavior. In this way, our method helps improve interaction with robots and language-assisted design systems, as well as assists artificial intelligence (AI) engineers in understanding and enhancing these models. To the best of our knowledge, we are the first to present and investigate counterfactual explanations for object referent identification. Our code is available at 
\mbox{\textit{https://github.com/toprei/why-are-you-wrong}}.
Our contributions include:
\begin{itemize}
    \item A novel method to generate valid and meaningful counterfactual utterances for object referent identification. 
    \item A comparison of multiple sampling strategies, that shows the superiority of an LLM-based sampler to generate counterfactual utterances.
    \item Evaluation with ShapeTalk data and three different models, that uncovers model bias and provides suggestions on how to improve the formulation of utterances.
    \item An LLM-based evaluation method to score semantic similarity between original and counterfactual utterances.
\end{itemize}

\label{chapter:introduction}

\section{Related Work}
\subsection{Object Referent Identification}
Object referent identification describes the task of selecting an object given a natural language utterance describing the object \cite{mitra2024whichone}. 
Specifically, we focus on blindfolded descriptions, i.e., utterances describing the object by its shapes and parts without using visual-attributes such as color or texture \cite{thomason2022snare}.
There exist several papers \cite{corona2022vlg, mitra2024whichone, sun2024da4lg}, which have developed models and methods for object referent identification.

Some works also provide datasets and benchmarks. ShapeGlot \cite{achlioptas2019shapeglot} introduces a dataset focused on chairs and lamps, with the task of selecting the correct object from three available options.
SNARE \cite{thomason2022snare} offers a benchmark dataset for grounding natural language expressions to distinguish 3D objects.
ShapeTalk \cite{achlioptas2023shapetalk} is with 536k utterances and 36k 3D models the largest and most recent dataset, which we use in our experiments.

\subsection{Counterfactual Explanations}
Post-hoc explanations of machine learning models help people to understand and act on algorithmic predictions \cite{mothilal2020dice}. A counterfactual explanation is an example-based XAI approach that explains a particular prediction outcome post-hoc, using a hypothetical example. 
This counterfactual example shows what could have been the outcome if an input to a model had been changed in a particular way \cite{verma2020counterfactual}.
In this way, counterfactual explanations do not explicitly answer the ``why” part of a decision, but also provides suggestions in order to achieve the desired outcome \cite{verma2020counterfactual}.
In this work, these suggestions take the form of natural language instructions, such as those for controlling a robot or interacting with a text-to-3D generative model.

Wachter et al.~\cite{wachter2017counterfactual} first introduced counterfactual explanations in 2017 as an optimization problem with two objective terms: one encouraging a class flip and the other minimal changes to the original data point. 
A popular choice for the optimization are genetic algorithms (GAs), which are founded upon the principle of evolution \cite{michalewicz2013genetic}. GAs allow a model-agnostic optimization for counterfactual examples without requiring access to the model internals \cite{sharma2020certifai}. 
Several works \cite{lash2017gic, sharma2020certifai, dandl2020moc, schleich2021geco} employ GAs to generate counterfactual examples. However, these methods primarily focus on tabular data and are not suitable for text. In this work, we study how GAs can be used to generate counterfactuals for object referent identification.

A few works have been dedicated to generate counterfactuals for text.
Li et al.~\cite{li2023prompting} created counterfactual examples for natural language understanding tasks by prompting large language models (LLMs). 
Similarly, Bhattacharjee et al. employed counterfactuals in two studies to stress-test natural language processing (NLP) models \cite{bhattacharjee2024zero} and to explain black-box text classifiers \cite{bhattacharjee2024towards}.
However, these prior works primarily focus on pure language tasks, which generally work with text consisting of multiple sentences \cite{kowsari2019text}. In contrast, geometric object referent identification works with geometric objects and a single carefully crafted utterance describing one object by its geometric attributes, making this task very sensitive to changes in individual words. 
In this work, we propose and investigate a method to generate counterfactual explanations for utterances in object referent identification.




\label{chapter:related_work}

\section{Method}
In this section, we first formally define the problem and then introduce our method for generating counterfactual utterances to explain misclassified samples. 
Our method generates first an initial set $P$ of counterfactual utterances using a specific sampling strategies. In a second step, the initial set is further optimized with a GA. Fig.~\ref{fig:overview} showcases a visual illustration of our method.

\subsection{Problem Definition}
We formally define an identification model $f(O, u)$ that takes two objects $O=(o_t,\:o_d)$, where $o_t$ and $o_d$ denote the target and distractor objects respectively, and the original utterance $u$ describing the target. In order to identify the target object based on the utterance, the model performs a binary classification $f(O,\:u)\:\mapsto [0, \:1]^2$. 
In the following $f(O,\:u)_t$ and $f(O,\:u)_d$ denote the predicted probabilities for the target and distractor objects respectively, summing up to 1, i.e., $f(O,\:u)_t + f(O,\:u)_d=1$.

A misidentification occurs when the model assigns to the distractor a higher probability than to the target object, i.e., $f(O,\:u)_t<f(O,\:u)_d$. In this case we want to find a counterfactual utterance $u'$ that leads to correct prediction $f(O,\:u')_t>f(O,\:u')_d$, while keeping a high semantic similarity $\text{Sim}(u,\: u')$ between the two utterances.




\subsection{Sampling}
Before starting to generate counterfactuals, it is necessary to identify which features are mutable and can be changed. A good counterfactual should not change any immutable features \cite{verma2020counterfactual}. 
However, textual utterances differ in structure and length, making it difficult to define features and their mutability in a traditional manner. 
To address this issue, selectively replacing certain types of words while retaining others help to preserve meaningful utterances. Thus, we restrict the mutable word types to content words, such as nouns, verbs, adjectives, and adverbs (NVAA), to maintain the core sentence structure while still allowing modifications to information-rich elements.
Previous works showed that models for object referent identification primarily rely on context-appropriate content words to successfully differentiate the target from the distractor \cite{achlioptas2019shapeglot}.
This advises to carefully craft counterfactual examples word by word.

In this paper, \textit{sampling} refers to the generation of a new utterance by the modification of a single word. To sample a counterfactual, a random NVAA word is selected from the original utterance and replaced by another word that is randomly picked from a word pool $W$. This word pool is a subset of all words in the available vocabulary. We investigate four sampling strategies that differ in how they define the word pool of potential replacement words:
\begin{enumerate}
    \item \textbf{Unaware:} Any NVAA word in the vocabulary.
    \item \textbf{Wordtype-aware:} Any NVAA word, but preserving the word type of the original to be replaced word.
    \item \textbf{Word-aware:} Synonyms to the original word.
    \item \textbf{Context-aware:} Words proposed by an LLM that takes into account the entire original sentence.
\end{enumerate}
For the context-aware variant, three state-of-the-art LLMs are used, Falcon3-1B-Instruct (Falcon) \cite{Falcon3}, Llama-3.2-1B-Instruct (LLaMA) \cite{dubey2024llama} and Qwen2.5-1.5B-Instruct (Qwen) \cite{qwen2.5}. 
These models are state-of-the-art, free to use, and among the most efficient in their model family yet still powerful enough to effectively respond to our prompt. 
After a random NVAA word is selected, we mask this word in the original utterance with the place holder ``WORD" and give the LLM the following prompt, where \{original word\} and \{masked utterance\} are replaced respectively.
\begin{quote}
  [Prompt]: \textit{Please replace in following sentence the word placeholder WORD with an actual word which is semantically similar to the word \{original word\}, but not the same word. Please give me for this five potential words. Here is the sentence: \{masked utterance\}.}
\end{quote}
Subsequently, the specific LLM name denotes the context-aware sampling strategy with the corresponding LLM sampler. 
Sampling from an utterance is repeated until an initial set of $N$ distinct counterfactuals is generated.

\subsection{Optimization}
After an initial set of $N$ counterfactual utterances is sampled, a GA is used to optimize this population. Similar to \cite{wachter2017counterfactual}, we define the counterfactual generation process as a bi-objective optimization problem.
Given a tuple of two objects $O$ and a counterfactual utterance $u'$, the first objective encourages a class flip from the distractor to the target object, defined as: 
\begin{multline} \label{eq:class_flip}
\text{Classflip}(O,\:u')  = \min(2\cdot f(O,\:u')_t,\:1) \\
+ 
\begin{cases}
-1 & \text{if $f(O,\:u')_t<f(O,\:u')_d$}, \\
0 &\text{otherwise.}
\end{cases}
\end{multline}
The second objective motivates the generation of an utterance that is semantically similar to the original, by maximizing the cosine similarity between the latent-representations given by the uniform sentence encoder (USE) \cite{cer2018universal}, a commonly used metric to measure the similarity between two sentences. This objective is defined as:
\begin{equation} \label{eq:similarity}
\begin{split}
\text{Sim}(u,\:u') = \frac{\text{USE}(u)\cdot \text{USE}(u')}{\|\text{USE}(u)\| \: \|\text{USE}(u')\|}
\end{split}
\end{equation}
The final fitness function is thus the unweighted sum of these two objectives:
\begin{equation} \label{eq:fitness}
\begin{split}
\text{Fitness}(O,\:u_,\:u') = \text{Classflip}(O,\:u') + \text{Sim}(u,\:u').
\end{split}
\end{equation}
It is worth mentioning that the first objective only rewards confidence gain until the class flip boundary. In this way, we only value whether a counterfactual is valid, i.e., flips the class, or not. In this manner, valid counterfactuals are compared solely on their similarity to the original utterance. Counterfactual examples should remain as close as possible to the original datapoint \cite{verma2020counterfactual}.
Moreover, a penalty of $-1$ is applied to utterances that are not valid, i.e., not flip the class. 
This encourages the GA to generate more valid counterfactuals by ensuring that a valid counterfactual with a low similarity score always has a higher fitness value than a non-valid counterfactual with a high similarity score.

The GA optimizes the population over $M$ generations by performing crossover and mutation. Mutation on a counterfactual utterance is performed by applying the chosen sampling function to change a single word. The GA generates multiple counterfactual utterances, however, we will always refer to a single counterfactual per sample, defined as the one with the highest cosine similarity, as in \eqref{eq:similarity}.

Algorithm \ref{alg:method} outlines our method in detail. The $\mathrm{Sample}$ function applies the previously described sampling by choosing a random NVAA word $w$ from the original utterance $u$ (line 2), creating then a word pool $W$ of potential replacement words (line 3) from which a randomly selected word $w'$ replaces the original word $w$ to create a counterfactual utterance $u'$ (line 4-5). This process is repeated until an initial set $P$ of $N$ distinct counterfactual utterances is generated (line 8-11). After that, the set $P$ is optimized with a GA for $M$ generations (line 12). In every generation, the fitness value of each utterance in $P$ is calculated (line 13), which is then used to select a mating pool $M$ to create crossover off-springs $C$ (line 14-15). 
Mutation is applied by randomly selecting off-springs with probability $p$ (line 16-17).
When an off-spring utterance is selected, it gets resampled by the chosen sampling strategy (line 18). In this way, the population $P$ gets iteratively updated (line 21).

\SetKwComment{Comment}{\#}{}
\SetKwInput{KwData}{Given}
\RestyleAlgo{ruled} 
\LinesNumbered

\begin{algorithm}
\caption{Counterfactual Utterance Generation}
\label{alg:method}
\SetAlgoLined
\KwData{model $f$, target and distractor objects $O$, original utterance $u$, sampling strategy $sampler$, population size $N$, number of generations $M$, mutation rate $p$}
\BlankLine
\SetKwFunction{FSample}{Sample}
\SetKwProg{Fn}{Function}{:}{end}
\Fn{\FSample{$sampler$, $u$}}{
    $w \gets \mathrm{randomNVAAword}(u)$\;
    $W = sampler(w,\:u)$\;
    $w' \gets \mathrm{randomChoice}(W)$\;
    $u' \gets \mathrm{replace}(w,\:w',\:u)$\;
    \Return $u'$\;
}
\BlankLine
\tcp{Sampling Initial Population}
$P \gets \{\}$\;
\While{$|P| < N$}{
    $P \gets P \cup \FSample(sampler,\:u)$;
} 
\BlankLine
\tcp{GA Optimization}
\For{$generation \gets 1$ \KwTo $M$}{
    $F \gets \mathrm{calculateFitness}(P,\:f,\:O,\:u)$\;
    $M \gets \mathrm{tournamentSelection}(P,\:F)$\;
    $C \gets \mathrm{singlePointCrossover}(M)$\;
    \tcp{Mutation}
    \For{$i \gets 1$ \KwTo $|C|$}{
        \If{$\mathrm{random} < p$}
        {
        $C[i] \gets \mathrm{\FSample}(sampler,\:C[i])$\;
        }       
        }
    $P \gets \mathrm{updatePopulation}(P,\:M,\:C)$\;
}
\Return $P$\;

\end{algorithm}


\label{chapter:method}

\section{Experiments}
In this section, we evaluate our approach for generating counterfactual utterances to explain misclassified samples in object referent identification.

\subsection{Experimental Setup}
The transformer-based neural listener from \cite{achlioptas2023shapetalk} serves as the to-be-explained object identifier. There exist multiple variants depending on the used backbone model. For our approach, we select three different models as shape backbones: a point cloud autoencoder (PC-AE) \cite{achlioptas2018pcae}, ResNet-101 (ResNet) \cite{he2016resnet}, and ViT-H/14/OpenCLIP (ViT) \cite{Radford2021vit}. 
These models cover different modalities, including point clouds and images, with ViT being state-of-the-art in the ShapeTalk \cite{achlioptas2023shapetalk} benchmark.

We select a set of 1\,000 misclassified samples from the ShapeTalk test set, which consists of 51\,234 samples in total, of which 3\,990 are misclassified by all three models.
From these, 1\,000 samples are randomly selected.

Since there is no prior work on counterfactual explanations for object referent identification, we use random search similar to \cite{dandl2020moc} as a baseline.
The baseline iteratively generates a pool of counterfactual utterances by randomly selecting an utterance from the pool and randomly replacing an NVAA word.
The pool initially contains only the original utterance.

For each model, we run experiments for all seven strategies, six versions of our method and the baseline. In every experiment of our method, we chose the initial population size $N$, akin to \cite{dandl2020moc}, to be 20, and the number of generations $M$ to be 30.
Therefore, the GA explores 620 counterfactual utterances per sample, which is also the budget for the baseline. The GA uses tournament-selection, single-point cross-over, and a mutation rate $p$ of 10 \%, balancing  exploration and exploitation.

\sisetup{separate-uncertainty=true,table-align-text-post=false}
\pgfplotstableread[col sep = comma]{"data/vit_random_search.csv"}\tablerandomsearch
\pgfplotstableread[col sep = comma]{"data/vit_unaware.csv"}\tableunaware
\pgfplotstableread[col sep = comma]{"data/vit_wordtype-aware.csv"}\tablewordtypeaware
\pgfplotstableread[col sep = comma]{"data/vit_word-aware.csv"}\tablewordaware
\pgfplotstableread[col sep = comma]{"data/vit_llama-1.csv"}\tablellama
\pgfplotstableread[col sep = comma]{"data/vit_falcon-1.csv"}\tablefalcon
\pgfplotstableread[col sep = comma]{"data/vit_qwen-1.5.csv"}\tableqwen
\begin{figure*}
\begin{center}
    \begin{tikzpicture}
        \begin{groupplot}[
        group style={
            group name=my plots,
            group size=2 by 1,
            xlabels at=edge bottom,
            ylabels at=edge left,
            horizontal sep=1.5 cm,
            vertical sep=1.5cm,
            },
        xlabel={Number of Generation},
        label style={font=\footnotesize},
        tick label style={font=\footnotesize},
        legend style={font=\footnotesize},
        legend cell align=left,
        legend style={
            at={(0.5,-0.5)}, 
            anchor=north,
            legend columns=4
        },
        xticklabel style={
            /pgf/number format/fixed,
            /pgf/number format/precision=4
        },
        scaled x ticks=false,
        width=\textwidth / 2,
        height=\textheight / 4,]

    \nextgroupplot[title=Success Rate, legend to name=plot_legend, ymin=0.6, ymax=1, ylabel=Success rate ratio]
    \addplot[name path=random-search, color=random-search, mark=x, mark size=1.5pt] table[x=generation, y=flip rate at least one] from {\tablerandomsearch};
    \addlegendentry{Random Search (baseline)};
    \addplot[name path=unaware, color=unaware, mark=x, mark size=1.5pt] table[x=generation, y=flip rate at least one] from {\tableunaware};
    \addlegendentry{Unaware};
    \addplot[name path=wordtype-aware, color=wordtype-aware, mark=x, mark size=1.5pt] table[x=generation, y=flip rate at least one] from {\tablewordtypeaware};
    \addlegendentry{Wordtype-aware};
    \addplot[name path=word-aware, color=word-aware, mark=x, mark size=1.5pt] table[x=generation, y=flip rate at least one] from {\tablewordaware};
    \addlegendentry{Word-aware};
    \addplot[name path=llama, color=llama, mark=x, mark size=1.5pt] table[x=generation, y=flip rate at least one] from {\tablellama};
    \addlegendentry{LLaMA};
    \addplot[name path=falcon, color=falcon, mark=x, mark size=1.5pt] table[x=generation, y=flip rate at least one] from {\tablefalcon};
    \addlegendentry{Falcon};
    \addplot[name path=qwen, color=qwen, mark=x, mark size=1.5pt] table[x=generation, y=flip rate at least one] from {\tableqwen};
    \addlegendentry{Qwen};

    \nextgroupplot[title=Latent Similarity, ymin=0.5, ymax=0.92, ylabel=Average cosine similarity]

    \addplot[name path=random-search, color=random-search, mark=x, mark size=1.5pt] table[x=generation, y=top-1 latent sim real cf per gen] from {\tablerandomsearch};
    \addplot[name path=unaware, color=unaware, mark=x, mark size=1.5pt] table[x=generation, y=top-1 latent sim real cf per gen] from {\tableunaware};
    \addplot[name path=wordtype-aware, color=wordtype-aware, mark=x, mark size=1.5pt] table[x=generation, y=top-1 latent sim real cf per gen] from {\tablewordtypeaware};
    \addplot[name path=Word-aware, color=word-aware, mark=x, mark size=1.5pt] table[x=generation, y=top-1 latent sim real cf per gen] from {\tablewordaware};
    \addplot[name path=llama, color=llama, mark=x, mark size=1.5pt] table[x=generation, y=top-1 latent sim real cf per gen] from {\tablellama};
    \addplot[name path=falcon, color=falcon, mark=x, mark size=1.5pt] table[x=generation, y=top-1 latent sim real cf per gen] from {\tablefalcon};
    \addplot[name path=qwen, color=qwen, mark=x, mark size=1.5pt] table[x=generation, y=top-1 latent sim real cf per gen] from {\tableqwen};

        \end{groupplot}
    \end{tikzpicture}
    \ref{plot_legend}
\caption[Plots]{Left: Success rate as ratio of samples for which at least one valid counterfactual utterance is generated. Right: Average latent cosine similarity of the best valid counterfactual per sample and generation.\label{fig:plot}}
\end{center}
\end{figure*}
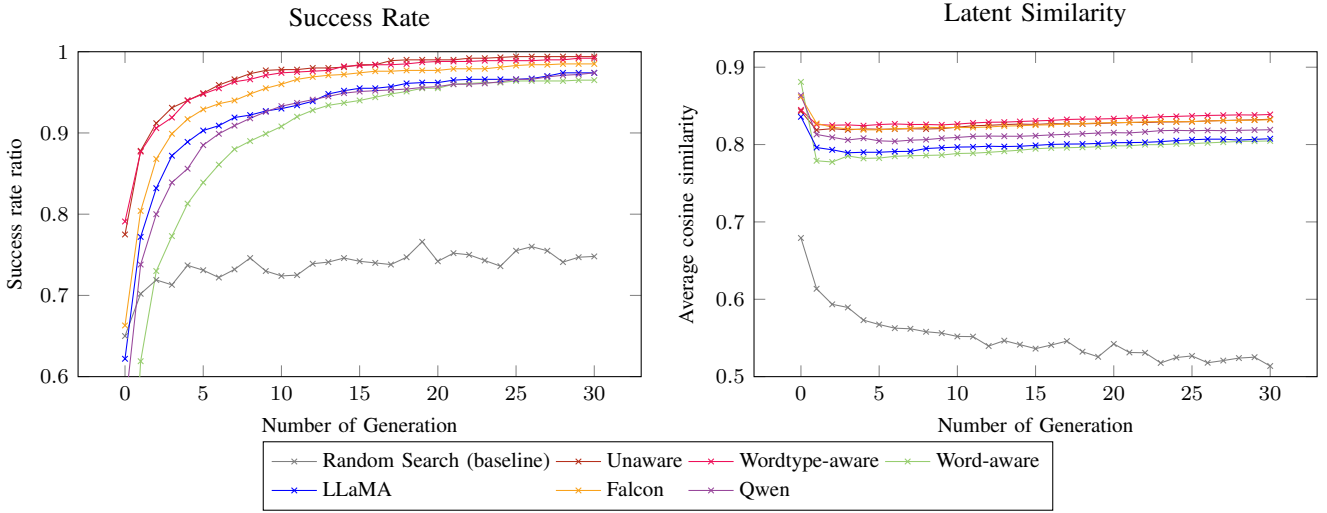

\subsection{Evaluation Metrics}
Since the primary goal is to generate valid counterfactuals, which are counterfactuals that are able to flip the class to the desired output \cite{verma2020counterfactual}, the success rate refers to the ratio of samples for which a strategy was able to produce at least one valid counterfactual.
A good counterfactual is edited only minimally in comparison to the original data point. The normalized Levenshtein distance \cite{levenshtein1966binary} allows to measure the edits in the token space. 
Text-based counterfactuals should not only be minimally edited, but also semantically similar to the original utterance.
Semantic similarity between two sentences can be measured by evaluating their embedding-based similarity. A popular choice is the cosine similarity of the USE \cite{cer2018universal} embeddings. This similarity score is also used in the optimization objective.

In addition to the embedding-based similarity, we evaluated the semantic adherence of the generated counterfactual to the original utterances, as well as their grammatical soundness, through an objective systematic evaluation.
Using a similarity evaluation that differs from the embedding-based similarity used during optimization allows for a more objective assessment.
The most common practice is to have humans score the similarity and grammatical correctness.
However, human evaluation is impractical for our analysis considering the quantity of entries. Alternatively, literature shows that a proper usage of LLMs as evaluators of generated linguistic contents achieves evaluation results that are acceptable to humans~\cite{chiang2023CanLargeLanguage,liu2023GEvalNLGEvaluation}. Therefore, we use the LLM GPT-4o-mini~\cite{openai2024gpt4o} to evaluate the quality of the counterfactuals produced by different strategies.
The pairwise analysis is conducted using the following prompts, beginning with the system message.
\begin{quote}
    [System]: \textit{You are an English language evaluation expert. You will then be given pairs of sentences denoted as reference sentences and candidate sentences. Each input is formatted as the reference sentence followed by the candidate sentence, with the delimiter ";".}
\end{quote}
The reference sentence represents the original utterance, while the candidate sentence denotes the counterfactual utterance that achieves the best fitness value during the optimization process. For each pair of references and candidate sentences, the LLM agent is first asked to analyze the grammatical correctness of the counterfactual utterance.
\begin{quote}
    [Grammar]: \textit{If the second sentence is grammatically incorrect, provide a brief explanation identifying the grammatical issues. Be tolerant of errors or anomalies caused by tokenization, stemming, cases, and lemmatization. Ignore errors in hyphenations of superlatives and comparatives. Keep the reasoning concise and focused only on the incorrectness.}
\end{quote}
It is further required to evaluate the semantic similarity between the original and the best counterfactual utterances.
\begin{quote}
    [Contextual]: \textit{Assess the level of semantic similarity between the candidate (second) sentence and the reference (first) sentence. Assign one of the following values, ranked from highest to lowest similarity: `equivalent', `very similar', `similar', `neutral', `dissimilar,' `very dissimilar', or `unrelated'. Respond only with the assigned value.}

    [Reason]: \textit{Provide a brief explanation highlighting the key elements that make the sentences semantically similar or dissimilar. Be tolerant, and keep the explanation concise and relevant.}
\end{quote}
Five rounds of repetitive yet independent evaluations are performed to minimize the impact of randomness in the results, with the final results determined through majority voting of LLMs responses across all five runs, i.e., by selecting the most frequently occurring response across all five runs.

\begin{table*}[hbt]
\centering
\caption{Quantitative comparison of different strategies for generating counterfactuals and the quality of the generated utterances. The \textbf{bolded} entries denote the best results. LD represents the normalized Levenshtein distance, and GC denotes the percentage of grammatically acceptable counterfactual utterances. 
The last six columns denote the degree of semantic similarities between all pairs of original and counterfactual utterances. 
More details can be found in section~\ref{chapter:experiments}.}
\label{table:quantitative}
\resizebox{\textwidth}{!}{%
    \begin{tabular}{lcccccccccccc}
    \toprule
    Model  &        Strategy & \thead{Success \\ Rate \% $\uparrow$} & \thead{Cosine \\ Similarity $\uparrow$} &  
    \thead{LD $\downarrow$} &
    \thead{GC \% $\uparrow$}
    & \thead{Equivalent \\ \% $\uparrow$}  &  \thead{Very \\ Similar \% $\uparrow$} & \thead{Similar \\ \% $\uparrow$ }& \thead{Neutral \\ \%}& \thead{Dissimilar \\ \% $\downarrow$} &  \thead{Very Dis-\\ similar \% $\downarrow$} \\
    \midrule
           & Random Search  & 99.1  & 0.786 & 0.215 & 18.9 & 0.3 & 12.8 & 4.6 & 0.7 & 75.5 & 6.1 \\
           & Unaware        & 99.9  & 0.851 & 0.203 & 15.8 & 0.2 & 13.7 & 3.6 & 1.3 & 75.8 & 5.4 \\
           & Wordtype-aware &  \textbf{100} & 0.852 &  \textbf{0.198} & 30.6 & 1.0 & 20.9 & 6.5 & 0.8 & 66.4 & 4.3 \\
    PC-AE  & Word-aware     & 99.7  & 0.823 & 0.256 & 19.0 & 3.4 & 25.8 & 4.8 & 1.8 & 59.8 & 4.4 \\
           & LLaMA          & 99.8  & 0.842 & 0.206 & 29.7 & 1.7 & 25.2 & 6.2 & 1.5 & 61.5 &  \textbf{3.9} \\
           & Falcon         & 99.9  &  \textbf{0.855} & 0.205 & 36.4 & 2.3 & 31.1 & 6.4 & 1.0 & 54.8 & 4.3 \\
           & Qwen           &  \textbf{100} & 0.846 & 0.215 &  \textbf{38.7} &  \textbf{4.6} &  \textbf{37.0} &  \textbf{7.4} & 0.9 &  \textbf{46.1} &  \textbf{3.9} \\ \midrule
           & Random Search  & 94.5  & 0.739 & 0.248 & 17.2 & 0.4 & 11.1 & 2.8 & 0.2 & 76.2 & 9.2 \\
           & Unaware        & 98.2  & 0.823 & 0.236 & 15.3 & 0.4 & 14.1 & 4.4 & 0.5 & 72.7 & 7.9 \\
           & Wordtype-aware &  \textbf{98.6}  &  \textbf{0.833} &  \textbf{0.216} & 28.7 & 0.4 & 21.2 & 5.2 & 1.4 & 66.8 &  \textbf{5.0} \\
    ResNet & Word-aware     & 96.7  & 0.797 & 0.296 & 17.9 & 1.8 & 20.3 & 5.3 & 1.0 & 64.6 & 7.0 \\
           & LLaMA          & 97.3  & 0.813 & 0.240  & 26.5 & 1.5 & 23.0 & 5.1 & 1.2 & 62.4 & 6.9 \\
           & Falcon         & 97.6  & 0.831 & 0.233 & 31.3 & 2.7 & 26.8 &  \textbf{6.0} & 0.8 & 58.6 &  \textbf{5.0} \\
           & Qwen           & 97.8  & 0.816 & 0.250  &  \textbf{32.0} &  \textbf{2.8} &  \textbf{28.9} & 5.9 & 1.3 &  \textbf{55.1} & 6.0 \\ \midrule
           & Random Search  & 94.6  & 0.738 & 0.25  & 18.2 & 0.4 & 11.2 & 3.2 & 0.7 & 75.7 & 8.8 \\
           & Unaware        &  \textbf{99.4}  & 0.833 & 0.223 & 13.9 & 0.1 & 14.7 & 3.6 & 0.7 & 75.4 &  \textbf{5.5} \\
           & Wordtype-aware & 99.2  &  \textbf{0.838} &  \textbf{0.211} & 31.5 & 0.4 & 23.4 & 4.4 & 0.7 & 64.6 & 6.4 \\
    ViT    & Word-aware     & 96.5  & 0.806 & 0.287 & 15.6 & 2.2 & 23.6 & 3.6 & 1.1 & 63.1 & 6.4 \\
           & LLaMA          & 97.4  & 0.808 & 0.246 & 25.7 & 2.0 & 22.3 & 5.9 & 1.1 & 61.9 & 6.8 \\
           & Falcon         & 98.5  & 0.832 & 0.239 & 31.6 & 2.0 & 26.4 & 4.8 & 1.0 & 60.1 & 5.6 \\
           & Qwen           & 97.4  & 0.819 & 0.254 &  \textbf{33.5} &  \textbf{3.7} &  \textbf{29.4} &  \textbf{6.2} & 1.3 &  \textbf{53.2} & 6.2 \\
           \bottomrule
    \end{tabular}
}
\end{table*}

\begin{figure*}[ht]
  \subcaptionbox*{Ratio of grammatically correct generated counterfactual utterances.}[.6\linewidth]{%
    \includegraphics[height=6.0cm,trim=0mm 0mm 0mm 0mm,clip]
    {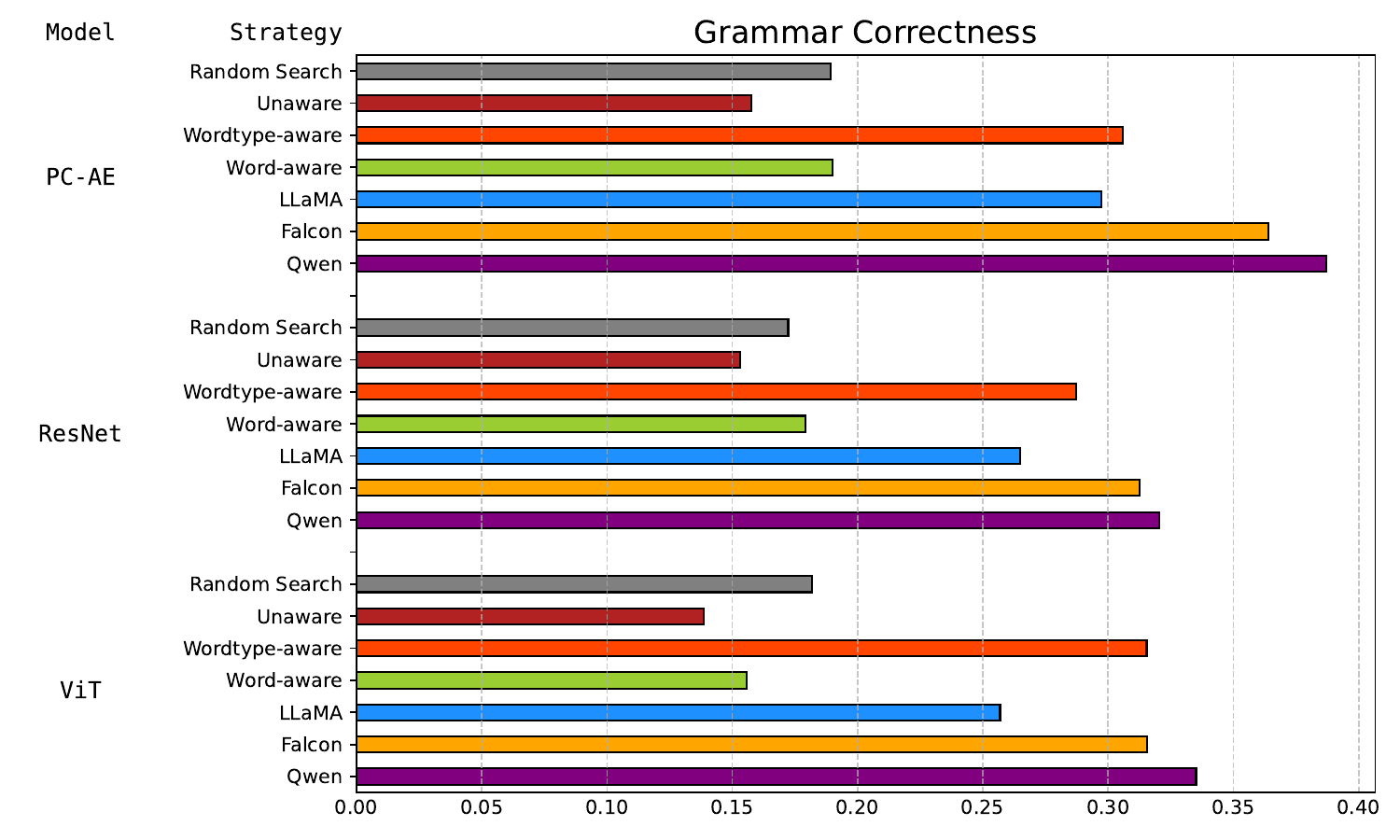}%
  }%
  \hfill
  \subcaptionbox*{Distribution of semantic similarity between the original and the counterfactual utterance.}[.4\linewidth]{%
    \includegraphics[height=6cm,trim=68mm 0mm 0mm 0mm,clip]
    {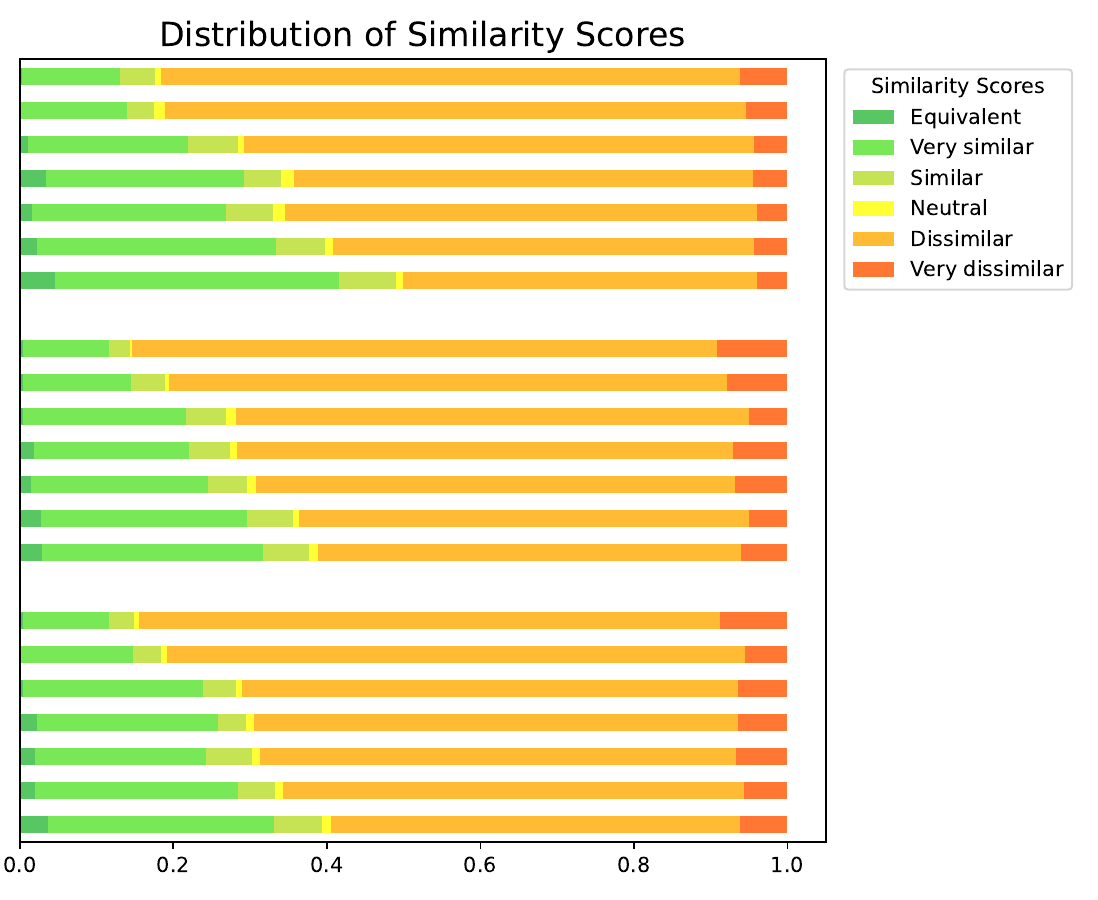}%
  }
  \caption{Majority vote results obtained from five repetitive but independent evaluation runs using GPT-4o-mini \cite{openai2024gpt4o} as a human surrogate~\cite{chiang2023CanLargeLanguage,liu2023GEvalNLGEvaluation} to analyze the grammatical correctness and the semantic similarity of the generated counterfactual utterances.}
  \label{fig:context_and_grammer}
\end{figure*}



\begin{table*}
\caption{Top-10 most frequent word pairs in 1\,000 counterfactuals generated by different strategies for the ViT model.}
\label{fig:top-10-pairs}
\centering
\begin{tabular}{ccc|ccc|ccc|ccc|ccc} 
\toprule
\multicolumn{3}{c}{Random Search} & \multicolumn{3}{c}{Unaware} & \multicolumn{3}{c}{Wordtype-aware} & \multicolumn{3}{c}{Word-aware} & \multicolumn{3}{c}{Qwen}  \\ 
\hline
replaced  & by & times & replaced  & by & times & replaced  & by & times & replaced  & by & times & replaced  & by & times \\
\hline
      has   &  distractor   & 2             & has  & not  &  19             &   has  & have   &  17             & has     &  have    & 87        &  has   & holds     &  11        \\
	 thick   &  hooped   & 2               & has  & is  &  15              &   thick  & thin   &  14           & legs    &  leg     & 36        & wide  & extensive &  9           \\
	 back   &  solid   & 2                 & has  & does  &  9             &   has  & is    &  14              & small   &  little  & 19        & has   & have      &  9             \\
	 has   &  doesnt   & 2                 & has  & it  &    5             &   has  & are    &  12             & wide    &  broad   & 15        & small & tiny      &  9             \\
	 has   &  no   & 2                     & back  & has  &  4	           &   thin  & thick     &  9          & short   &  little  & 11        & legs  & leg       &  8                \\
	 side   &  too   & 2                   & small  & is  &  4 	           &   are  & has   &  7               & as      &  equally & 9         & wide  & large     &  7               \\    
	 base  &  strap   & 1                  & thin  & large  &  4	       &   have      & has  &  7           & thin    &  lean    & 9         & thin  & frail     &  7             \\
	 are    &  cylinder   & 1              & long  & short  &  4 	       &   small      & short  &  6        & have    &  get     & 9         & have  & carry     &  6         \\
	 shade   &  knight   & 1               & legs  & not  &  3  	       &   long   & thick  &  5            & back    &  rear    & 8         & has   & hold      &  6          \\
      target   &  size   & 1                & has  & no  &  3               &   tall  & short   &  5            & drawers &  drawer  & 7         & short & small     &  6             \\
\bottomrule
\end{tabular}
\end{table*}

\begin{figure*}[ht]
\begin{center}
  \subcaptionbox*{Replaced words}[.5\textwidth]{%
    \adjincludegraphics[height=4.1 cm,trim={{.2\width} {.1\width} {.2\width} {.1\width}},clip]{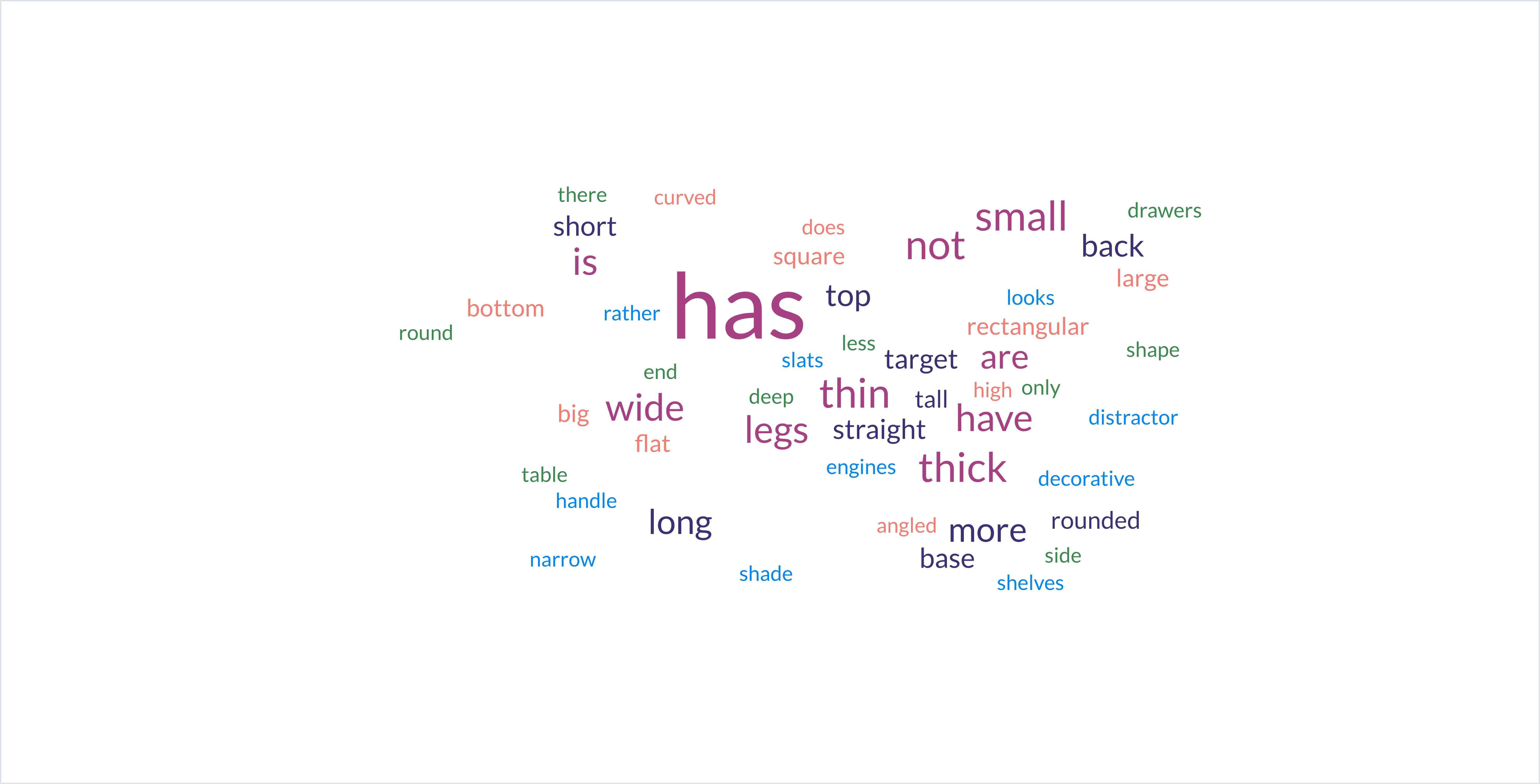}
  }%
  \hfill
  \subcaptionbox*{Inserted words}[.5\textwidth]{%
    \adjincludegraphics[height=4.1 cm,trim={{.1\width} {.1\width} {.2\width} {.1\width}},clip]{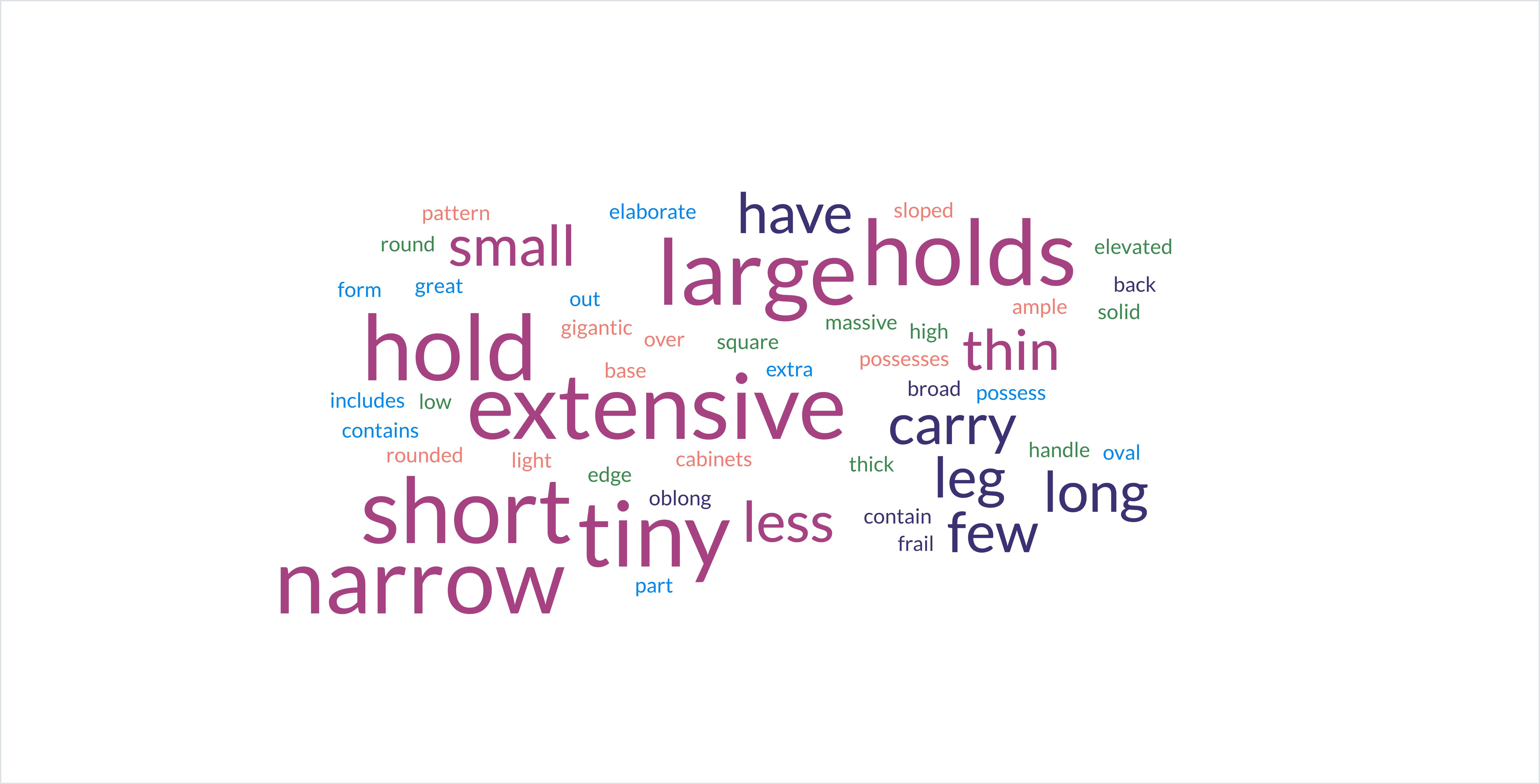}
  }
  \caption{Commonly replaced and inserted words by the Qwen sampler to generate counterfactuals for the ViT model.}\label{fig:word_cloud}
\end{center}
\end{figure*}

\begin{figure*}
\centering
{\footnotesize
\begin{tabularx}{\textwidth}{@{}cccc@{}}
 & Distractor\phantom{wwwwww}Target & Distractor\phantom{wwwwww}Target & Distractor\phantom{wwwwww}Target  \\
 & \adjincludegraphics[height=2.2cm,trim={{.1\width} {.13\width} {.1\width} {.1\width}},clip]{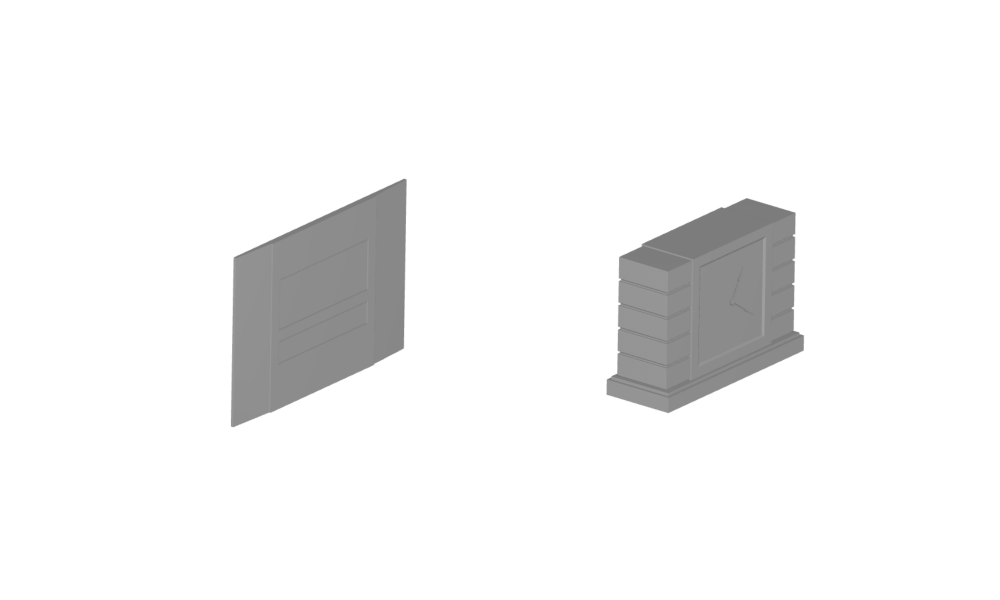} 
 & \adjincludegraphics[height=2.2cm,trim={{.1\width} {.13\width} {.1\width} {.1\width}},clip]{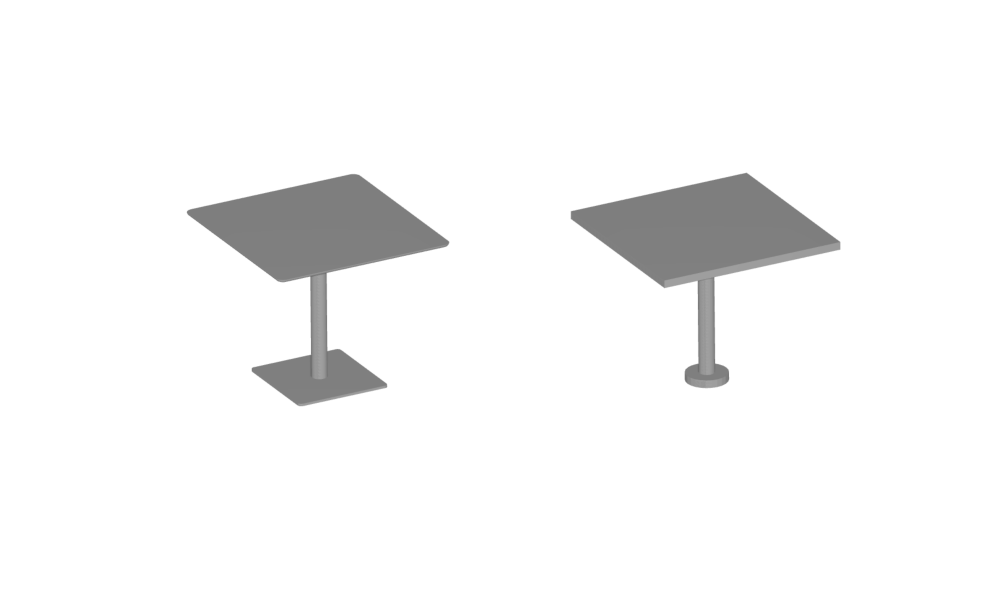} 
 & \adjincludegraphics[height=2.2cm,trim={{.1\width} {.13\width} {.1\width} {.1\width}},clip]{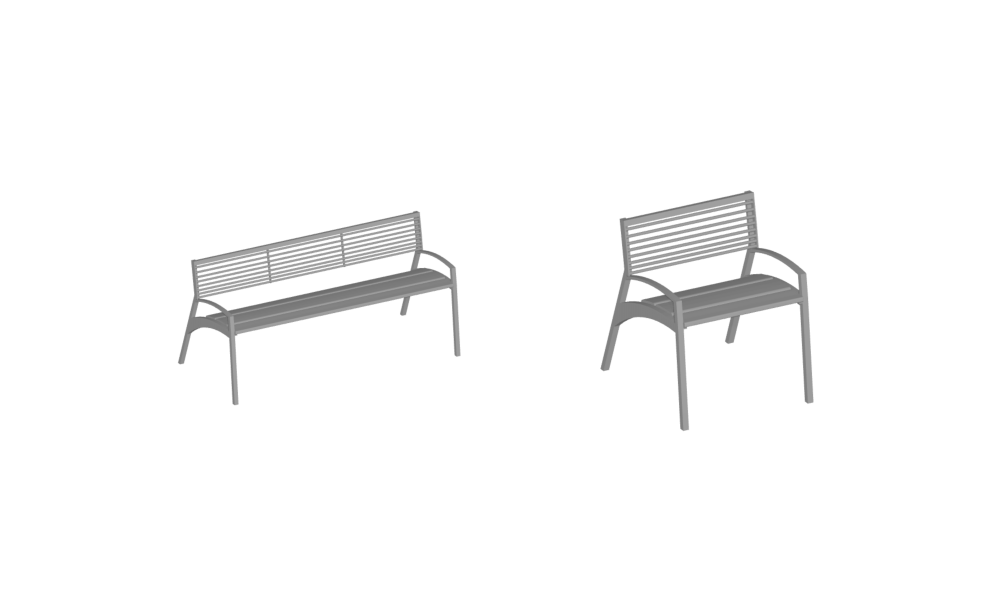} \\
original  & it looks like it is \textcolor{replaced}{\textbf{made}} of bricks & the top does not \textcolor{replaced}{\textbf{have}} rounded corners & the back \textcolor{replaced}{\textbf{has}} no vertical supports behind it \\
counterfactual  & it looks like it is \textcolor{inserted}{\textbf{constructed}} of bricks & the top does not \textcolor{inserted}{\textbf{possess}} rounded corners & the back \textcolor{inserted}{\textbf{hold}} no vertical supports behind it \\

 & \adjincludegraphics[height=2.2cm,trim={{.1\width} {.13\width} {.1\width} {.1\width}},clip]{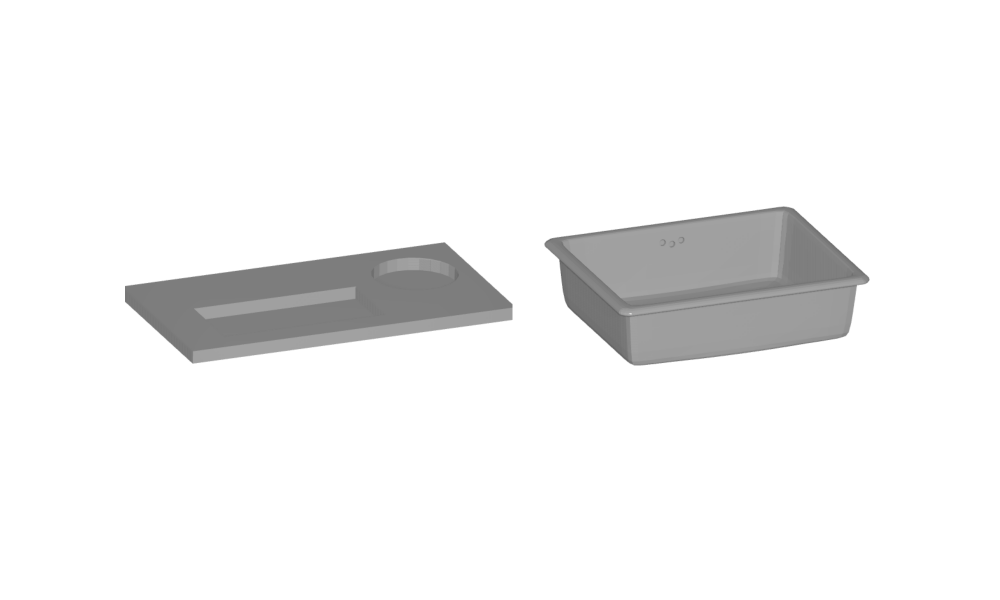}
 & \adjincludegraphics[height=2.2cm,trim={{.1\width} {.13\width} {.1\width} {.1\width}},clip]{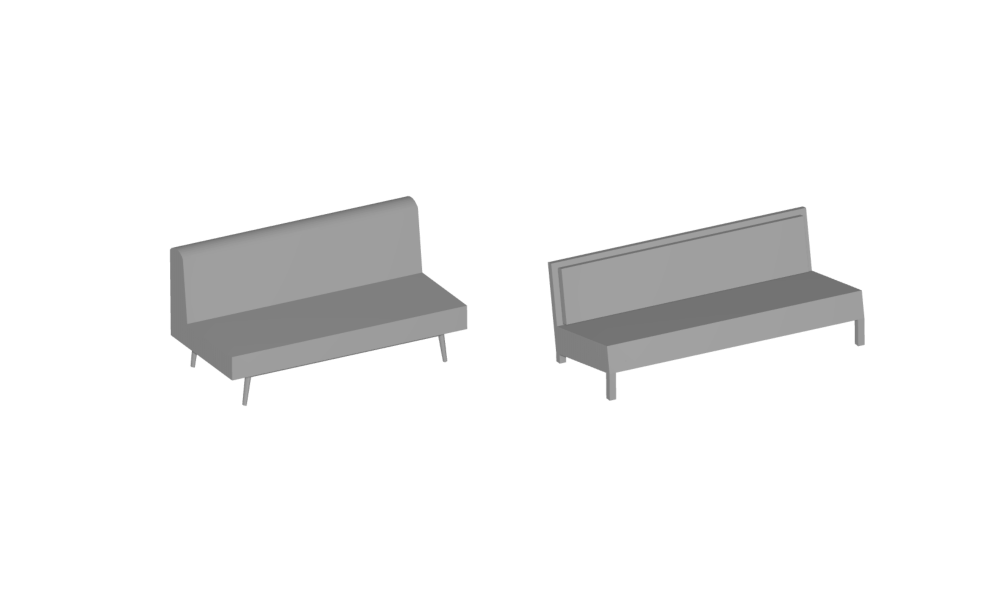}
 & \adjincludegraphics[height=2.2cm,trim={{.1\width} {.13\width} {.1\width} {.1\width}},clip]{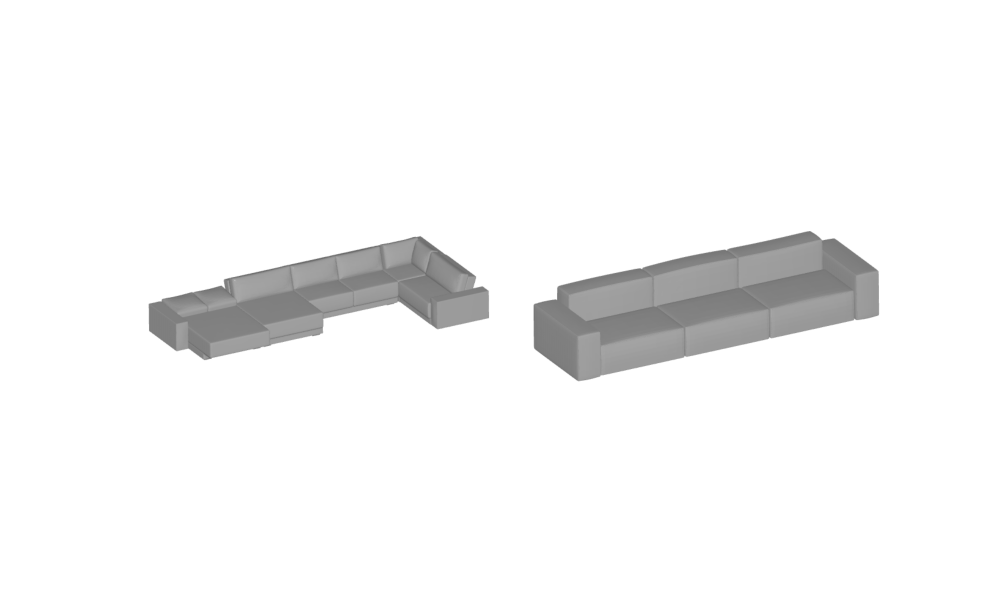} \\
original  & there are three \textcolor{replaced}{\textbf{places}} for knobs & its back rest has sharper \textcolor{replaced}{\textbf{edges}} & the target has three \textcolor{replaced}{\textbf{unique}} seating \textcolor{replaced}{\textbf{areas}} \\
counterfactual  & there are three \textcolor{inserted}{\textbf{spots}} for knobs & its back rest has sharper \textcolor{inserted}{\textbf{ridges}} & the target has three \textcolor{inserted}{\textbf{distinctive}} seating \textcolor{inserted}{\textbf{locations}} \\

 & \adjincludegraphics[height=2.2cm,trim={{.1\width} {.13\width} {.1\width} {.1\width}},clip]{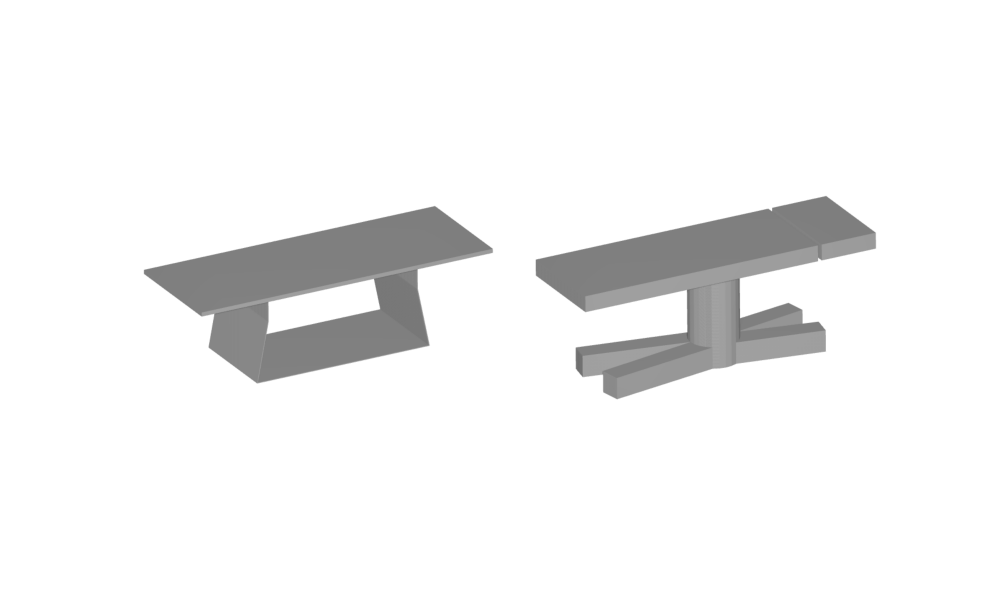}
 & \adjincludegraphics[height=2.2cm,trim={{.1\width} {.13\width} {.1\width} {.1\width}},clip]{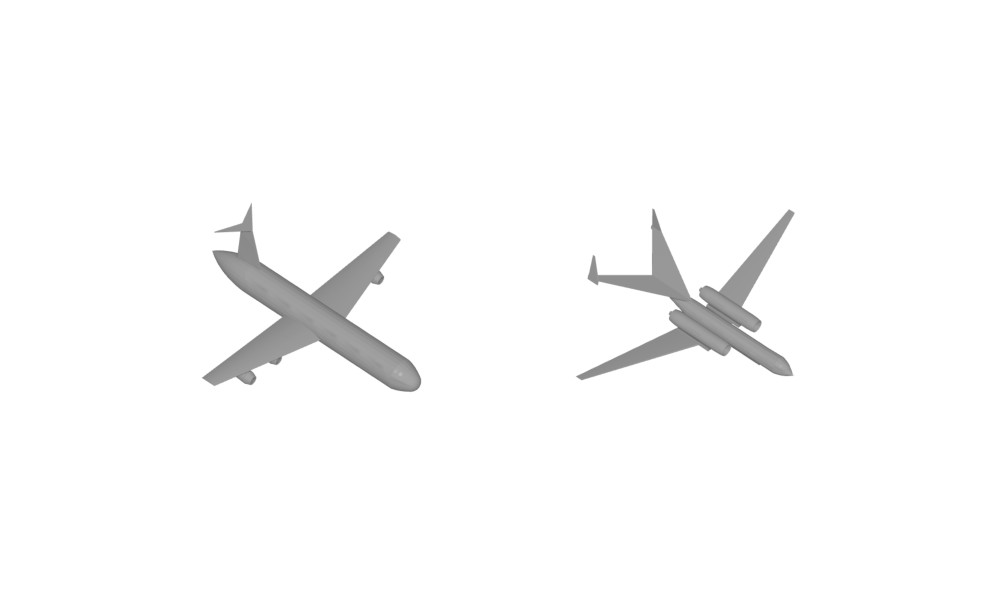}
 & \adjincludegraphics[height=2.2cm,trim={{.1\width} {.095\width} {.1\width} {.1\width}},clip]{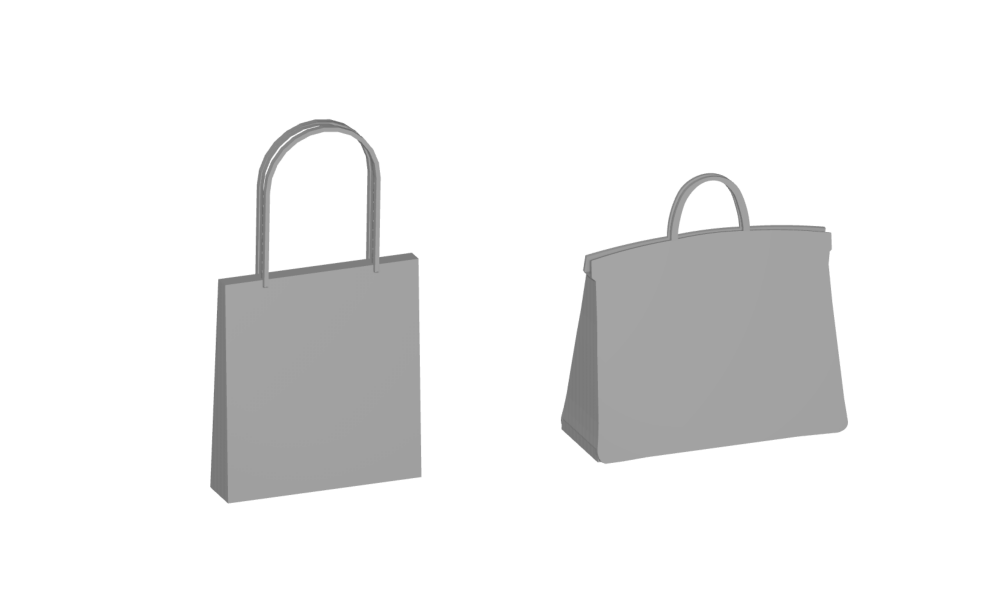} \\
original  & the legs are not \textcolor{replaced}{\textbf{curved}} & there are two \textcolor{replaced}{\textbf{large}} engines & its handle is much \textcolor{replaced}{\textbf{smaller}} \\
counterfactual  & the legs are not \textcolor{inserted}{\textbf{bent}} & there are two \textcolor{inserted}{\textbf{huge}} engines & its handle is much \textcolor{inserted}{\textbf{shorter}} \\

 & \adjincludegraphics[height=2.2cm,trim={{.1\width} {.11\width} {.1\width} {.1\width}},clip]{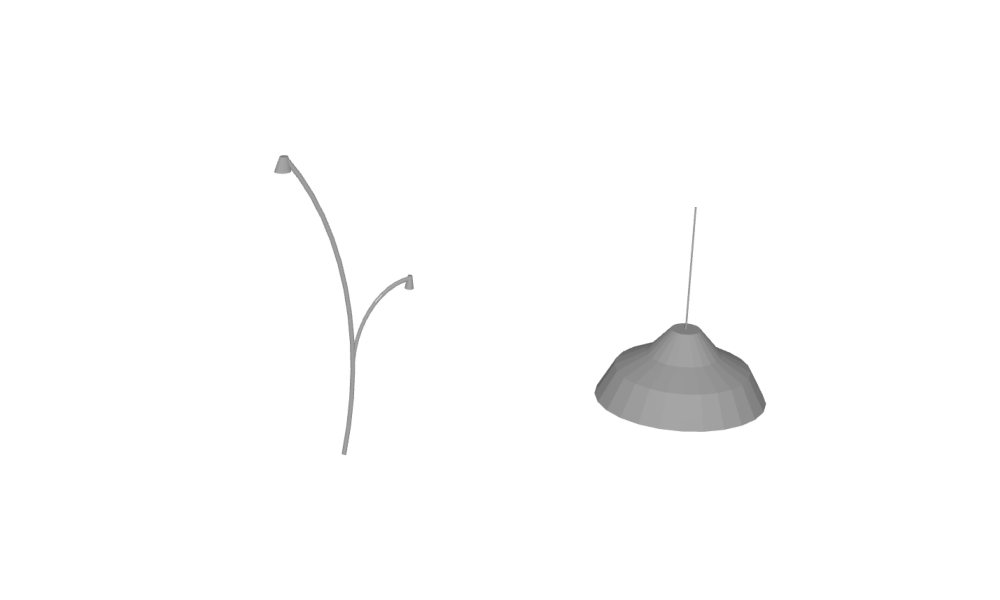}
 & \adjincludegraphics[height=2.2cm,trim={{.1\width} {.1\width} {.1\width} {.1\width}},clip]{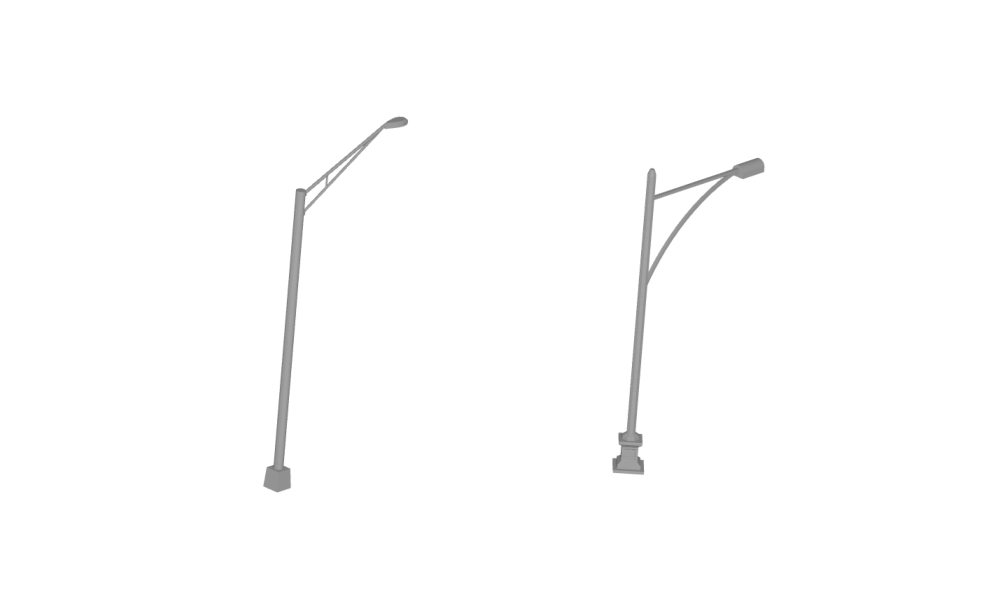}
 & \adjincludegraphics[height=2.2cm,trim={{.1\width} {.13\width} {.1\width} {.1\width}},clip]{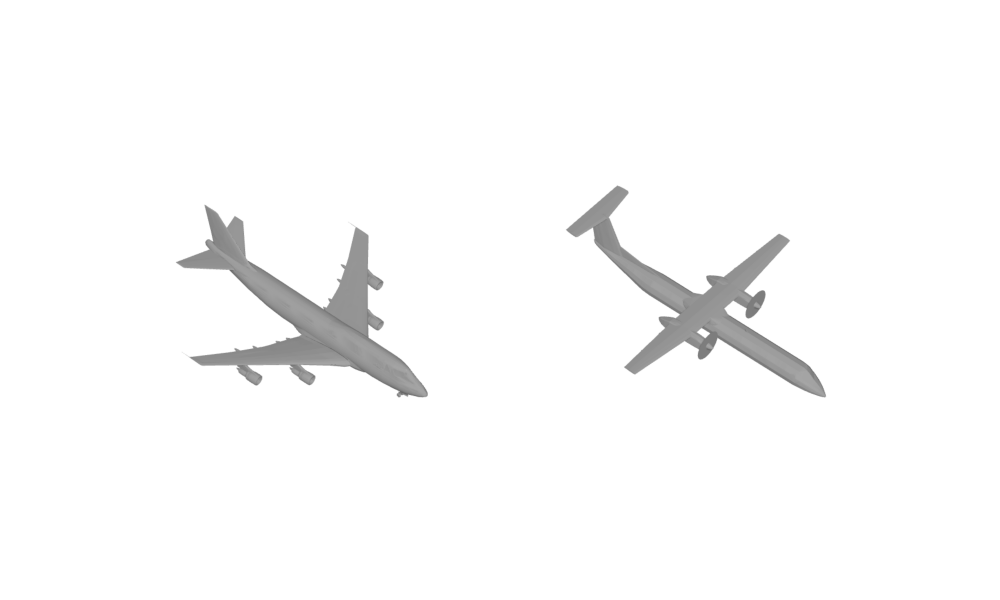} \\
original  & it is \textcolor{replaced}{\textbf{hanging}} & its light is \textcolor{replaced}{\textbf{rectangular}} & the target has two \textcolor{replaced}{\textbf{engines}} under the wings \\
counterfactual  & it is \textcolor{inserted}{\textbf{droop}} & its light is \textcolor{inserted}{\textbf{oblong}} & the target has two \textcolor{inserted}{\textbf{propellers}} under the wings \\

 & \adjincludegraphics[height=2.2cm,trim={{.1\width} {.13\width} {.1\width} {.1\width}},clip]{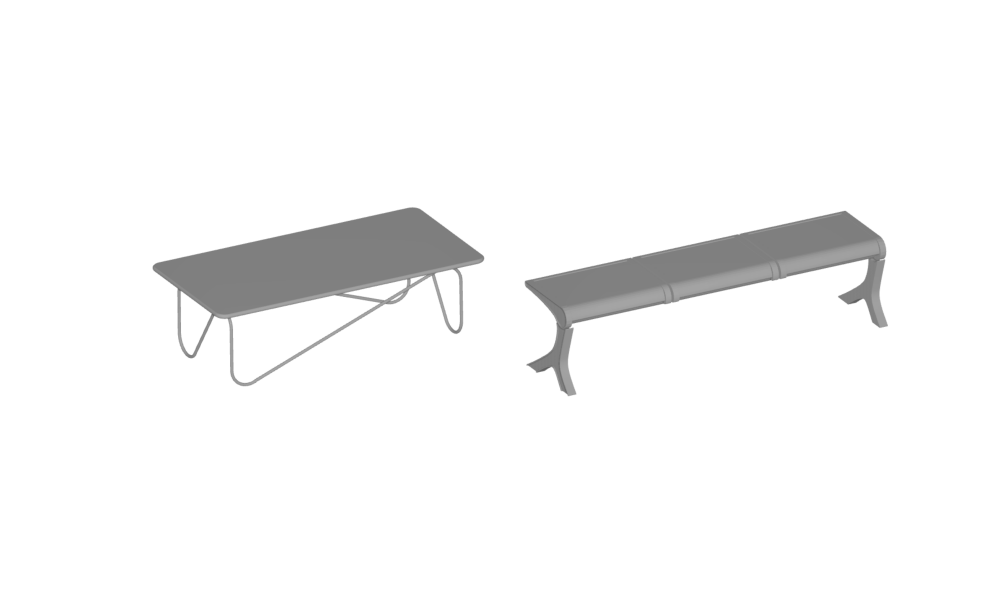}
 & \adjincludegraphics[height=2.2cm,trim={{.1\width} {.13\width} {.1\width} {.1\width}},clip]{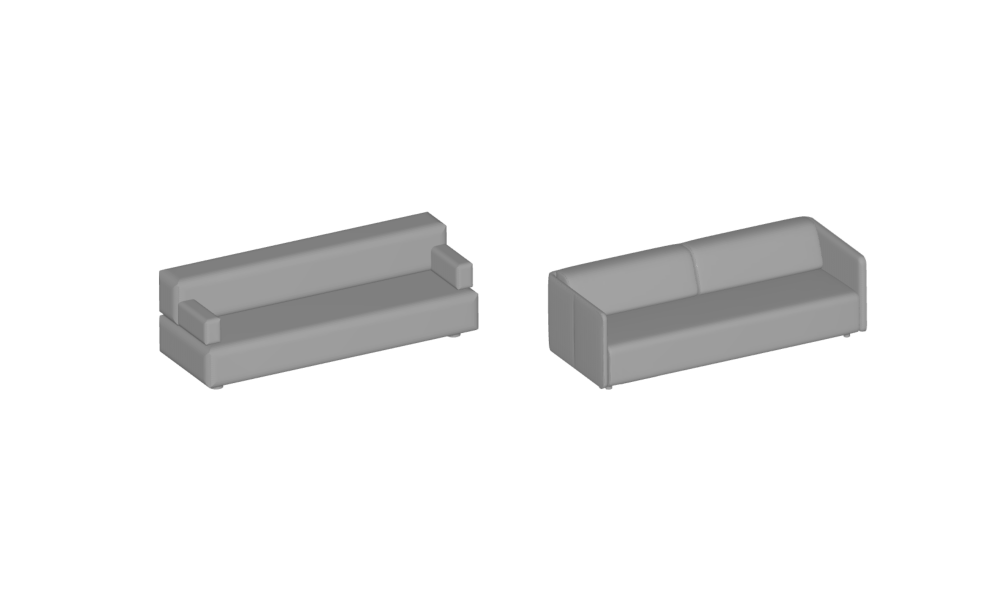}
 & \adjincludegraphics[height=2.2cm,trim={{.1\width} {.13\width} {.1\width} {.1\width}},clip]{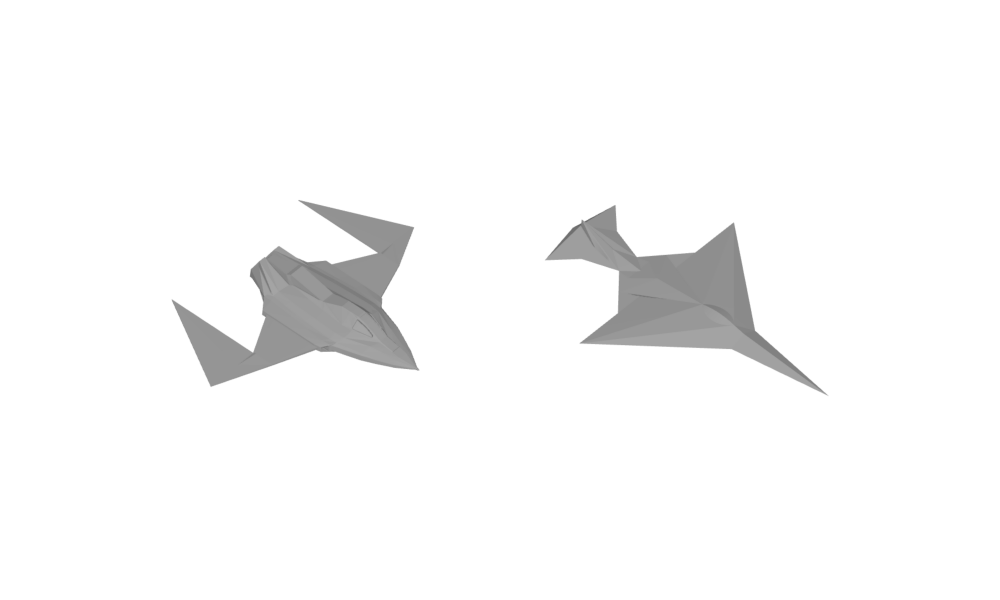} \\
original  & it has \textcolor{replaced}{\textbf{curved}} \textcolor{replaced}{\textbf{legs}} & there are \textcolor{replaced}{\textbf{no}} regular arm rests & the tail is bulky and \textcolor{replaced}{\textbf{wide}} \\
counterfactual  & it has \textcolor{inserted}{\textbf{smooth}} \textcolor{inserted}{\textbf{knees}} & there are \textcolor{inserted}{\textbf{merely}} regular arm rests & the tail is bulky and \textcolor{inserted}{\textbf{narrow}} \\
\end{tabularx}
}
\caption{Qualitative results of our method. Given the original utterance, the model misclassifies the left object as the target. With our counterfactual utterances, the model correctly identifies the target object on the right. Our counterfactuals maintain the structure of the original utterance, are semantically similar, meaningful, and provide insights why a prediction was wrong. All counterfactuals were generated using Qwen as sampler and ViT as model.
}\label{fig:qualitative}
\end{figure*}

\subsection{Results}
\textbf{Optimization.} The left plot in Fig.~\ref{fig:plot} demonstrates that the optimization effectively increases the success rate, i.e., the ratio of samples in which at least one valid counterfactual utterance is generated. In comparison, random search has a significantly lower success rate per generation, i.e., per $N$ newly generated counterfactuals.

The right plot, shows that the GA in our method also increases the average latent similarity per generation after an initial drop. The drop in latent similarity after the first generation is a result of the cross-over function. Generation zero contains the initially sampled population with only one word changed compared to the original utterance. However, after the first generation, the cross-over function produces utterances with more than one word changed. This causes the average similarity to first drop before it begins to rise again slightly.
Conversely, for random search, the average latent similarity per generation declines. Over time, the pool of candidate utterances used for resampling contains a growing number of utterances that differ increasingly from the original. 

\textbf{Quantitative evaluation.} Table \ref{table:quantitative} demonstrates that our method consistently achieves a high success rate in generating valid counterfactuals. The random search baseline performs slightly less effectively than our proposed method. A similar trend can be observed for the average cosine similarity. This shows that the optimization in our method effectively improves both objectives.
It is worth noting that the disparities in success rates across the models can be attributed to the differing modalities. Point clouds may capture less visual details compared to images, which makes it easier to generate counterfactual utterances that are capable of convincing PC-AE to flip the class.
The normalized Levenshtein distance indicates a reasonable small editing. Only about 20 to 25 percent of words in an utterance are changed for most strategies.

However, a significant difference can be seen in grammatical correctness and in the distribution of similarity scores presented in Table \ref{table:quantitative} or visualized in Fig.~\ref{fig:context_and_grammer}. 
The context-aware LLM-samplers LLaMA, Falcon, and Qwen generate counterfactual utterances more similar to the original compared to any other sampling strategy. 
Only the PC-AE-LLaMA combination falls marginally behind the word-aware strategy. 
Furthermore, the grammatical correctness of the utterances generated by the LLM-samplers is notably superior, with the word-type aware sampling strategy being the only significant competitor.
One possible explanation for this behavior is that the word-type aware sampling strategy, which only allows replacing words with those of the same type, effectively prevents the generation of grammatically incorrect utterances.

\textbf{Frequent word pairs.} Table \ref{fig:top-10-pairs} shows the most frequent word pairs. The random search baseline is not capable of producing reasonable commonly used pairs. In contrast, the unaware and wordtype-aware strategies produce more commonly used word pairs, but with limited semantic plausibility. 

Both the word-aware and the context-aware LLM-sampler Qwen are able to produce meaningful replacement pairs. However, the synonym-based word-aware strategy generates much less diverse pairs and a large portion is occupied by the two pairs ``has-have" and ``legs-leg", which are not informative for explaining a misclassified sample and might lead to grammatically incorrect utterances. 
The ratio of grammatically correct utterances produced by the word-aware strategy is significantly lower than that of Qwen, as shown in Fig. \ref{fig:context_and_grammer}.

\textbf{Commonly replaced and inserted words:} Fig.~\ref{fig:word_cloud} visualizes in two word clouds the individual words that are commonly replaced and inserted. The larger a word appears, the more frequently it is replaced or inserted. The figure refers to the words used by the Qwen sampler for generating counterfactuals for the ViT model. 

The word ``has" is the most frequently replaced word by a significant margin.
A reason for this might be that this word occurs regularly in the original utterances. The inserted words are more diverse, but mainly consist of adjectives and verbs, indicating that predictions can be enhanced by refining these rather than using nouns.

\textbf{Qualitative evaluation.} 
Fig.~\ref{fig:qualitative} shows counterfactuals generated by our method along with the original utterance and the two objects. 
Given the original utterance, the model misidentifies the distractor as the target object, whereas our counterfactuals lead to a correct identification.
The red words depict the replaced words, while green is for the inserted ones.

These examples qualitatively demonstrate the ability of our method to explain false-identified samples. 
Exemplary are the words ``small" and ``short". Our findings indicate that a model tends to associate ``small" more with a lesser horizontal dimension in relation to the vertical dimension, while ``short" is specifically linked to height. Examples of this specific case are shown in Fig.~\ref{fig:overview} and Fig.~\ref{fig:qualitative}.

Furthermore, many counterfactuals show that employing stronger adjectives, such as ``huge", ``bent", or ``oblong", help to identify the target object.
Also, using more precise verbs and nouns is beneficial, as illustrated in the first and second row of Fig.~\ref{fig:qualitative}. 
Our method also identifies subtle errors in the formulation, such as using ``engines" instead of ``propellers" in the context of an airplane.
Moreover, it can also generate more creative counterfactuals, which must be carefully considered within the context of the two objects. Examples are depicted in the last row of Fig.~\ref{fig:qualitative}.


\label{chapter:experiments}


\section{Conclusions}
In this paper, we presented a method to generate counterfactual explanations for object referent identification.
We empirically demonstrated that our approach is capable of generating valid and meaningful counterfactual utterances, which help explain why a model made an incorrect prediction by revealing weaknesses in the original utterance and model bias towards specific words.
These counterfactuals provide practitioners with suggestions on how to formulate utterances to, for example, control a robot or interact with a 3D generative model. Furthermore, our method assists AI engineers in understanding and improving models for object referent identification.
In particular, we showed that the context-aware variant of our method is superior in producing contextual similar and grammatical correct counterfactuals utterances.

A current limitation of our method lies in the embedding-based similarity measurement during optimization. This metric under-penalize the insertion of contextually unrelated words as long as semantical key parts of the utterance are retained.

In future work, we plan to explore alternative similarity metrics and scoring functions.
Additionally, our work on generating counterfactual utterances could be used for data augmentation, potentially enhancing the performance of state-of-the-art models for object referent identification.

\label{chapter:conclusion}

\FloatBarrier
\printbibliography 


\end{document}